\documentclass[a4paper]{article}

\usepackage[margin=1in]{geometry} % full-width

\usepackage{amsmath}
\usepackage{amsthm}
\usepackage{amssymb}

\usepackage[utf8]{inputenc}
\usepackage{hyperref}
\hypersetup{
	unicode,
	pdfauthor={Joshua Goldman, John K. Tsotsos},
	pdftitle={Statistical Problems with Image Datasets},
	pdfsubject={Dataset Bias},
	pdfkeywords={Datasets, Bias, Computer Vision},
	pdfproducer={LaTeX},
	pdfcreator={pdflatex}
}

\usepackage{biblatex} %Imports biblatex package
\theoremstyle{plain}

\theoremstyle{definition}

\usepackage{graphicx, color}
\graphicspath{{fig/}}

\usepackage{changepage}
\newenvironment{bias-ex}{\begin{adjustwidth}{1cm}{}}{\end{adjustwidth}}

\usepackage{makecell}
\usepackage{algorithm, algpseudocode} % use algorithm and algorithmicx for typesetting algorithms
\usepackage{mathrsfs} % for \mathscr command

\usepackage{lipsum}

\title{Statistical Challenges with Dataset Construction: Why You Will Never Have Enough Images}
\author{
    Josh Goldman$^1$ \\% \thanks{...}
    \and
    John K. Tsotsos$^2$
}

\date{
	$^1$Dept. of Mathematics, McGill University, Montreal, Canada \\ \texttt{joshua.goldman2@mail.mcgill.ca}\\%
    $ $ \\
	$^2$Dept. of Electrical Engineering and Computer Science, York University, Toronto, Canada \\ \texttt{tsotsos@yorku.ca}\\[2ex]%
}

\begin{document}
\maketitle
	
\begin{abstract}
    Deep neural networks have achieved impressive performance on many computer vision benchmarks in recent years. However, can we be confident that impressive performance on benchmarks will translate to strong performance in real-world environments? Many environments in the real world are safety critical, and even slight model failures can be catastrophic. Therefore, it is crucial to test models rigorously before deployment. We argue, through both statistical theory and empirical evidence, that selecting representative image datasets for testing a model is likely implausible in many domains. Furthermore, performance statistics calculated with non-representative image datasets are highly unreliable. As a consequence, we cannot guarantee that models which perform well on withheld test images will also perform well in the real world. Creating larger and larger datasets will not help, and bias aware datasets cannot solve this problem either. Ultimately, there is little statistical foundation for evaluating models using withheld test sets. We recommend that future evaluation methodologies focus on assessing a model's decision-making process, rather than metrics such as accuracy.
\end{abstract}

\section{Introduction} \label{introduction}
Whether explicit or implicit, every real-world domain has an acceptable error rate \cite{tsotsosProbingEffectSelection2021}. Determining whether a product is sufficiently safe for deployment in a given domain is challenging. Different fields have developed their own standards that technologies must satisfy before deployment in the real world. Much work has analyzed why users do and do not trust AI systems \cite{yangUserTrustArtificial2022, jermutus2022influences, glikson2020human, asan2020artificial}. Some factors are related to the system itself, such as predictability \cite{nordheimInitialModelTrust2019} and functionality \cite{bedueCanWeTrust2021}. Others tend to the psychological side, such as the quality of the interface design \cite{ameenCustomerExperiencesAge2021}. However, trust in a model is only warranted if the trust was caused by the model's ability to perform a given functionality \cite{jacoviFormalizingTrustArtificial2021}. If a model acquires trust through factors unrelated to its functionality, such as a sleek user interface, the model has acquired unwarranted trust. Although an attractive interface is certainly correlated with high functionality, it does not cause high functionality. Rather, the two are both caused by an underlying source, such as a large project budget or talented developers. AI models can demonstrate warranted trustworthiness both intrinsically and extrinsically \cite{jacoviFormalizingTrustArtificial2021}. 

\begin{enumerate}
    \item \textbf{Intrinsic trust} is trust in the model's decision-making process. A model is \textbf{intrinsically trustworthy} if its decision making process matches our assumptions for how that process should be. For instance, a model might demonstrate trustworthiness if its reasoning for detecting a banana was the yellow colour and the curvy shape. A model might demonstrate untrustworthiness if its reasoning for detecting a ship is the presence of water, not any features of the ship itself. This definition is subjective, since different individuals possess different prior assumptions for what is a reasonable decision making process.
    
    \item \textbf{Extrinsic trust} is trust in the process used to test the model. A model is \textbf{extrinsically trustworthy} if it was assessed with a rigorous testing regimen that sufficiently covered the target domain. 
\end{enumerate}

Presently, evaluations of computer vision safety and performance heavily rely on extrinsic assessment. Current methods for testing models in computer vision rely on a withheld test set which is used at the end of training to assess model accuracy. However, many unanswered questions surround the testing process. How many images should the test set contain? How should the testing images be collected? Are there any guarantees that a model which performs well on the test set will also perform well in production? Is there any way to analyze the probability that a model will generalize to unseen images? To the best of our knowledge, rigorous answers to the above questions do not exist. As deep neural networks (DNNs) are increasingly used in many safety-critical domains, such as autonomous vehicles \cite{raoDeepLearningSelfdriving2018} and medical image diagnosis \cite{panayidesAIMedicalImaging2020}, the development of a rigorous methodology for assessing the performance of computer vision models is ever more critical. 

Our paper begins by exploring safety standards and requirements in the real world. We demonstrate that acceptable failure rates are extremely low, much lower than the error rates attained by many of today's state-of-the-art models. We then present a list of potential sources of bias that can, and likely do, impact image datasets, followed by a survey of the current dataset creation practices. We discuss empirical work demonstrating the bias in current computer vision datasets, as well as the difficulty DNNs face when tested with out-of-distribution data. Then, we review the theory of survey sampling, and demonstrate the flaws with current methods to assess a model's performance and error rate. Finally, we analyze other error assessment methods and illustrate the challenges that arise when applying them to computer vision. 

Our primary contributions are as follows:
\begin{enumerate}
    \item We provide, to the best of our knowledge, the first argument grounded in non-random sampling theory that disputes the usability of withheld test sets.

    \item We demonstrate that dataset bias is an unsolvable problem. Regardless of how many factors a dataset curator explicitly controls for, the final dataset will still differ greatly from that of a random sample. Future work on dataset bias will benefit from knowledge of this limitation.
\end{enumerate}

\section{Safety Requirements in the Real World} \label{safety-real-world}
Safety requirements in real-world environments are stringent. The safety integrity level (SIL) of a part is its required reliability, measured in failures per hour for parts in continuous operation (High Demand) and failures per use for infrequently used parts (Low Demand). The SIL of a part is determined through a risk analysis. High-risk parts are categorized as SIL 4, low-risk parts are categorized as SIL 1 \cite{redmill2000understanding}. The maximum allowable failure rates for Low Demand and High Demand parts rated at different SILs are shown in Table \ref{tab:sil_table}. Safety integrity levels were developed by the International Electrotechnical Commission (IEC) \cite{bell2006introduction}, a world leader in electrical standards supported by nearly 170 countries.

\begin{table}[h]
    \centering
    \caption{Acceptable failure rates for different SIL Levels \cite{bell2006introduction}}.
    \def\arraystretch{1.5}
    \begin{tabular}{| c | c | c |}
    \hline
        SIL & \makecell{Low Demand \\ (Failure Per Use)} & \makecell{High Demand \\(Failure Per Hour)} \\ 
        \hline
        $4$ & $10^{-4}$ & $10^{-8}$ \\
        \hline
        $3$ & $10^{-3}$ & $10^{-7}$ \\
        \hline
        $2$ & $10^{-2}$ & $10^{-6}$ \\
        \hline
        $1$ & $10^{-1}$ & $10^{-5}$ \\
        \hline 
    \end{tabular}
    \label{tab:sil_table}
\end{table}

Many industries also have their own maximum tolerable failure rates. The target probability of failure on aircrafts is $10^{-7}$ failures per flight hour, and each key component must have a failure rate below $10^{-9}$ failures per flight hour \cite{federalaviationadministrationGeneralFunctionInstallation1996}. In many drug studies, researchers will only accept the results if the probability of a false positive is less than $0.001$ \cite{serdarSampleSizePower2021}. Road vehicles should fail no more than $10^{-8}$ times per driving hour \cite{chenBinFIEfficientFault2019}. The UK's maximum tolerable risk of fatality for workers in large-scale industrial plants is $10^{-3}$ fatalities per year, and the target rate is $10^{-6}$ fatalities per year \cite{healthandsafetyexecutiveGuidanceALARPDecisions, muhlbauer15RiskManagement2004}. The probability of a nuclear reactor failure that could kill more than $100$ people should be below $10^{-7}$ failures per year in the UK \cite{officefornuclearregulationONRSafetyAssessment2020}.

State-of-the-art ImageNet classification algorithms scored approximately $99\%$ top-$5$ accuracy \cite{byerlyCurrentStateArt2022}, yielding an error rate around $0.01$. If an algorithm with similar performance is used in a self-driving car, countless mis-classifications will occur. Many of these failures may not be catastrophic, and additional subsystems in the car could catch the error. However, like any component in a safety-critical system, an image classifier will have an associated maximum failure rate. An image classifier that fails once every $10^{8}$ images is still unusable in a car with an acceptable failure rate of once every $10^{9}$ images. Our point here is that that performance alone is not relevant. One must compare performance against real world safety targets to evaluate the effectiveness of a model \cite{tsotsosProbingEffectSelection2021}. Therefore, rigorous performance estimates with small error bounds are essential in the real world.

\section{Formulating the Problem} \label{formulating-problem}
For a specified computer vision task, we define the target population as the set of all possible images that could possibly be passed to the model for inference. We denote the target population as $T$ and the number of images in the target population as $N$. Next, we define the true model accuracy, $\bar{p}$, which is the model's average accuracy across every single image in the target population. More precisely, for each image $t_i$ in $T$ where $1 \leq i \leq N$, let $p_i$ be the model's accuracy on the image $t_i$. Then, 
\begin{equation*}
   \bar{p} = \frac{1}{N}\sum p_i 
\end{equation*}
For instance, if we were measuring binary accuracy, then $p_i$ would be $1$ if the model correctly classified the image and $0$ otherwise. We also note that a weighted average can be taken instead, if certain images are more likely to occur or if failures on certain images may lead to more catastrophic outcomes.

As an example, consider the task of classifying common household objects in $1200 \times 1200$ RGB images. Each image has $1200 \times 1200 \approx 1,000,000$ pixels, and each pixel has $3$ channels which can take any value from $0$ to $255$. Thus, there are a total of 
\begin{equation*}
    256^{1200 \times 1200 \times 3} \geq 100^{3,000,000} = 10^{6,000,000}
\end{equation*}
images. Many of these images will appear as random noise to a human observer. Thus, $T$ is much smaller since we only consider images that could be captured by a camera. A very conservative lower bound on the number of $1200 \times 1200$ images that could possibly be captured by a camera is $10^{400}$ \cite{pavlidisNumberAllPossible2009}. The specific computer vision task you are pursuing will also further reduce the number of possible images. However, "whether the number of all possible images is $10^{30}$ or $10^{3000}$...either number implies that it is impossible to conduct research that relies on 'knowing' all possible images." \cite{pavlidisNumberAllPossible2009}

The impossibility of 'knowing' all possible images becomes a challenge when calculating $\bar{p}$. Calculating the precise value of $\bar{p}$ requires testing on every image in $T$ which is completely infeasible. Thus, we must test the model on a sample of images from the target population, and, based on the results, estimate the value of $\bar{p}$. Following standard statistical practices, our estimate for $\bar{p}$ should also be paired with a confidence interval, indicating our uncertainty in the estimate \cite{simStatisticalInferenceConfidence1999}. For instance, after testing a DNN one might estimate that the accuracy of the model on an image classification task is $93.5 \pm 0.2 \%$. Confidence intervals are not routinely reported in deep learning papers at the present moment, though confidence intervals for DNNs are an active research area \cite{kuleshov2018accurate,tagasovska2019single,cortes2018deep}.

\section{Possible Sources of Bias} \label{sources-bias}
Collecting images for a test sample is a difficult activity that must be handled with care. The number of biases that can plague a dataset are immense. Much work has been done cataloguing the sources of bias in research. Sackett meticulously identified $35$ potential sources of bias in analytic research \cite{sackettBiasAnalyticResearch1979}. The Catalogue of Bias extends Sackett's efforts, listing numerous sources of bias that can impact research studies \cite{nunanCatalogueBias2017}. We outline some sources of bias related to the data collection process, and assess their relevance to image datasets \cite{tsotsosProbingEffectSelection2021}.

\vspace{0.2cm}

\noindent \textbf{Apprehension Bias} occurs ``when a study participant responds differently due to being observed" \cite{brasseyApprehensionBias2019}.

\begin{bias-ex}
    People may pose and look differently when they know that they are being photographed. Hence, a model may perform poorly if it is trained on images of people who are aware that they are being photographed, but tested on images of people who are unaware that they are being photographed. 
\end{bias-ex}

\noindent \textbf{Ascertainment Bias} is when there are ``systematic differences in the identification of individuals included in a study or distortion in the collection of data in a study" \cite{spencerAscertainmentBias2017}.

\begin{bias-ex}
    Different subjects, light conditions, locations, camera settings, and other features may be more likely to be included in the dataset than others. Performance may drop when the model is tested on images with features that were less likely to be included in the training set. For instance, sunny images may have a higher likelihood of being included in the dataset, causing the model to perform poorly on darkly lit images.
\end{bias-ex}

\noindent \textbf{Availability Bias} is ``a distortion that arises from the use of information which is most readily available, rather than that which is necessarily most representative" \cite{banerjeeAvailabilityBias2019}.

\begin{bias-ex}
    The images from the top results of an internet search are convenient and easy to collect. However, they may not represent the entire image class.
\end{bias-ex}

\noindent \textbf{Chronological Bias} occurs ``when study participants allocated earlier to an intervention or a group are subject to different exposures or are at a different risk from participants who are recruited later" \cite{spencerChronologicalBias2017}.

\begin{bias-ex}
    The distribution of images in a given domain may change over time. A model that may have performed well on images sampled from one time period may struggle on images from a previous or subsequent time period. 
\end{bias-ex}

\noindent \textbf{Volunteer bias} occurs when ``participants volunteering to take part in a study intrinsically have different characteristics from the general population of interest" \cite{mahtaniVolunteerBias2017}.

\begin{bias-ex}
    The individuals that volunteer to take photos for datasets may differ from the population in many features, such as geography and socio-economic status.
\end{bias-ex}

\noindent \textbf{Wrong sample size bias} occurs ``when the wrong sample size is used in a study: small sample sizes often lead to chance findings, while large sample sizes are often statistically significant but not clinically relevant" \cite{spencerWrongSampleSize2017}.

\begin{bias-ex}
    While machine learning datasets seem large, are they large enough to reflect the complexity of the visual world? Is it even possible to reflect the complexity of the visual world in a human-created dataset?
\end{bias-ex}

\vspace{0.2cm}

\noindent Additional sources of bias in visual datasets have also been analyzed \cite{fabbrizziSurveyBiasVisual2022, mehrabiSurveyBiasFairness2021}, such as
\begin{enumerate}
    \item \textbf{Negative set bias} is the under-representation of a negative of a class in a dataset. For example, not-boat in a dataset of images at sea \cite{torralbaUnbiasedLookDataset2011}.
    \item \textbf{Availability bias} occurs when dataset creators use the most convenient data sources, such as search engines and pre-existing datasets that may have been developed for other purposes. 
    \item \textbf{Geographical bias.} refers to biases in the imbalance of geographic locations at which images were taken \cite{shankarNoClassificationRepresentation2017}.
    \item \textbf{Correlation bias} refers to a correlation between two groups in a dataset that falsely or harmfully connects them together. For instance, if all photographs of women were taken in a kitchen.
    \item \textbf{Crawling bias} refers to biases of the web crawling and scraping algorithm that is used to collect photos. 
    \item \textbf{Platform bias} refers to biases due to the platform from which the data is collected (for example, Google, X, Facebook, etc.).
    \item \textbf{Capture bias} refers to biases in how the image is structured, such as positioning, lighting, background, and camera settings \cite{torralbaUnbiasedLookDataset2011}.
\end{enumerate}
The list above is not meant to be comprehensive. The list's purpose is to show the broad range of biases that can affect an image dataset. It is highly improbable that one could create a high-quality dataset that is balanced along the above listed biases, without carefully planning the dataset creation process. Indeed, in the following sections, we illustrate that arbitrarily collecting data, the process used to construct many modern image datasets, will almost certainly yield biased results. 

\section{Current Datasets} \label{current-datasets}
Throughout the history of image datasets, there has been an effort to minimize bias and produce higher quality representations of our environments. From Lena, to COIL, to Caltech-101, to PASCAL-VOC, to ImageNet, each dataset attempted to outdo the complexity of its predecessor and better capture the nuances of the visual world \cite{torralbaUnbiasedLookDataset2011}. Some datasets were collected manually with a camera \cite{nayarColumbiaObjectImage1996}, some were scraped from the internet \cite{dengImageNetLargescaleHierarchical2009}, and some used a combination of both techniques \cite{winnObjectCategorizationLearned2005}. 

However, much evidence has shown that image datasets are a biased representation of the visual world \cite{ponceDatasetIssuesObject2006, pintoWhyRealWorldVisual2008, yangFairerDatasetsFiltering2020}. Machine learning models absorb the biases present in the training set and sometimes even amplify them \cite{zhao2017men}. Furthermore, the performance of a model on a biased test set typically overestimates the model's true performance $\bar{p}$. Image classifiers perform substantially worse when tested on images from other datasets \cite{torralbaUnbiasedLookDataset2011}. Models trained to classify gender performed much better on lighter, male faces than on darker, female faces \cite{buolamwiniGenderShadesIntersectional2018}. Classifiers trained on ImageNet and CIFAR10 performed much worse when tested on new test sets assembled with the same data collection process as the original datasets \cite{rechtImageNetClassifiersGeneralize2019}. One explanation for why models perform better on biased test sets is that models trained on biased data can perform well by learning decision shortcuts. These shortcuts may work for biased data, but they do not transfer to the real world \cite{geirhosShortcutLearningDeep2020}. Hence, we need to be certain that our model has not learned spurious correlations from the training set and that it can generalize to the entire target population within some acceptable error. The examples mentioned above corroborate the claim that withheld test sets are an ineffective method for rejecting the possibility that a model learned decision shortcuts. 

In light of these findings, many researchers created bias-aware datasets. The Pilot Parliaments Benchmark controls for skin type and gender in a facial images dataset \cite{buolamwiniGenderShadesIntersectional2018}. The BDD100k dataset collected driving videos from four locations and ensured a wide representation of different weather conditions \cite{yuBDD100KDiverseDriving2020}. The PIQ23 dataset is a dataset of portraits, controlling for numerous conditions such as lighting, age, skin tone, gender, and subject-to-lens distance \cite{chahineImageQualityAssessment2023}. For a comprehensive review of bias aware datasets, see \cite{fabbrizziSurveyBiasVisual2022}. Recent work has also provided guidelines and suggestions for dataset creators, such as recommendations for building human centric datasets \cite{andrewsPrinciplismGuidedResponsible2023}, templates for documenting dataset creation \cite{gebru2021datasheets}, and advice for sociocultural data collection and annotation \cite{joLessonsArchivesStrategies2020}.

While the specific details of data collection differ for every dataset, the fundamental principles remain the same:
\begin{enumerate}
    \item The dataset creators specify a collection of classes. The classes can be simple, such as cats and dogs, or they can be more detailed, such as images of red cars on sunny days. Regardless of how detailed or simple the class descriptions are, there are still a finite set of image classes. 
    \item The dataset creators will also specify the number of images they want in each class. This step is sometimes skipped.
    \item Images are collected for each class using search engines, manual photography, or a combination of both. 
    \item The collected images are annotated and sorted into their corresponding class. Annotation methodologies and their problems are beyond the scope of this paper. 
\end{enumerate}
Are these data collection methods adequate for assessing a model? Are the test sets collected representative? There was a prevalent assumption that large-scale datasets were an accurate representation of the visual world because they contained millions of images \cite{pavlidisNumberAllPossible2009}. However, is this a sufficient condition for representation? If the target population consists of $10^{20}$ images and we sample $1$ billion images, we have sampled $0.00000001 \%$ of the total number of images. Is that enough?

\section{Sampling Theory} \label{sampling-theory}
Sampling theory deals with the challenge of selecting a sample from a population \cite{lohrSamplingDesignAnalysis2021}. There are two main methods of sampling: random sampling and non-random sampling \cite{acharyaSamplingWhyHow2013}. In random sampling, also commonly known as probability sampling, each unit in the population has a non-zero probability of being selected for the sample, and units are selected for the sample using a randomized process. In non-random sampling, often called non-probability sampling, units are not selected randomly. Instead, another sampling mechanism is used, such as convenience sampling, purposive sampling, or quota sampling \cite{lohrSamplingDesignAnalysis2021}. 

Random sampling is a  mathematically rigorous framework for making generalizations to a larger population from a sample. One can be confident that inferences from a random sample are accurate and that error estimates are inversely proportional to the sample size \cite{lohrSamplingDesignAnalysis2021, wuSamplingTheoryPractice2020}. Non-random sampling, on the other hand, does not have the same guarantees \cite{lohrSamplingDesignAnalysis2021}. The non-random sampling technique used to collect current image datasets is called \textit{quota sampling}. Quota sampling is a non-random sampling method where the target population is divided up into subgroups. Within each subgroup, units are sampled according to a non-random mechanism \cite{lohrSamplingDesignAnalysis2021}. Of the currently known non-random sampling methods, effective quota sampling is assumed to yield the most accurate results \cite{yangQuotaSamplingAlternative2014}.

In theory, quota samples only yield reliable estimates for the variable of interest when the following four assumptions hold \cite{changbaowuStatisticalInferenceNonprobability2022}.
\begin{enumerate}
    \item The variables used to define the quota groups characterize the participation probabilities of the units.
    \item The selection mechanism is relatively random within each quota group.
    \item The size of each quota is known from a reliable external source. 
    \item Perennial non-respondents and respondents take on similar values of the target variable. 
\end{enumerate}
Proving these assumptions is challenging and must often rely on unattainable information. For instance, how can one prove that perennial non-respondents behave like respondents, when perennial non-respondents are impossible to survey. Similarly, as we show later in the paper, precisely determining the level of randomness in a non-random sample is difficult.

In practice, quota sampling yields poor estimates compared to high-quality random samples. In Neyman's famous paper, he demonstrated that a non-random sample of towns in Italy yielded unsatisfactory results for the uncontrolled variables \cite{neymanTwoDifferentAspects1934}. Initial research into quota sampling showed that random samples and well-executed quota samples yielded similar results \cite{moserExperimentalStudyQuota1953}. However, these results were questionable, since the random sample had a low response rate. Subsequent studies demonstrated that the results of a quota sample and a random sample deviated as the response rate of the random sample increased \cite{yangQuotaSamplingAlternative2014}. Significant differences emerged when the response rate of the random sample was high. Additional research has corroborated these findings \cite{cummingProbabilitySamplingAlways1990, dutwinApplesOrangesGala2017, macinnisAccuracyMeasurementsProbability2018, pennayOnlinePanelsBenchmarking2018}. The conclusion we draw here is that quota sampling, despite being superior to other non-random sampling methods, still fails to adequately represent the population.

\section{When Can We Collect a Random Sample?} \label{random-sample}
In general, collecting a random sample of images is challenging. Enumerating every image in the target population and choosing a random collection of them is likely impossible. However, in some domains there exists a data generating process which, if a set of assumptions hold, would imply the randomness of a collected sample.

For instance, one can collect every single X-Ray image captured in a specific hospital over the course of a year \cite{homeyerRecommendationsCompilingTest2022}. If one assumes that X-Ray images do not change much yearly, then testing on the collection of all images taken that year can serve as a random sample of images. However, it should be noted that, under the current assumptions, this collection of images is only a random sample of images at the particular hospital it was collected from. If one were to treat this collection as a random sample of images for all the hospitals in an area, they would have to include the additional assumption that X-Rays at nearby hospitals follow the same distribution. In general, the larger the domain, the more assumptions are needed to imply the randomness of the sample. This is a challenge, since the randomness of a sample is only as strong as the assumptions it rests on. If one assumption is invalid, then the implied randomness of the test sample no longer holds. 

Testing a model on a random sample will yield accurate measurements of the model's true abilities, provided that one tests the model on enough data to yield sufficiently small uncertainties. Further research is needed to determine in which domains it is possible to collect a random sample. However, based on the current dataset collection practices discussed in Section \ref{current-datasets}, it is clear that most modern datasets are not random samples. Hence, as of right now, random samples are the exception, not the rule. Therefore, the rest of this paper focuses on the vast majority of domains where collecting a random sample of images is impossible, or impractical at the present moment. The next section explores the accuracy of error estimates based on non-random samples. 

\section{Statistical Error Estimates in Computer Vision} \label{error-vision}
Turning back to computer vision, we remember that model accuracy, $\bar{p}$, is the parameter of interest we are estimating. For a sample of images $\hat{T}$ from $T$, let $n$ denote the size of $\hat{T}$ and $\hat{p}$ denote the model's average accuracy on the images in $\hat{T}$.

Denote the sampling mechanism of $\hat{T}$ as $R$. Thus, $R_{i}=1$ if the image $t_i$ is included in the sample, and $R_{i}=0$ if $t_i$ is not included. In a random sample of images from the target population where each image has an equal probability of being chosen, the correlation between the sampling mechanism and $p$ is $0$ \cite{mengStatisticalParadisesParadoxes2018, lohrSamplingDesignAnalysis2021}. That is, the random sampling mechanism is not intentionally choosing challenging or easy images for the model. However, based on the empirical results discussed in Section \ref{current-datasets}, there appears to be a correlation between the current dataset sampling mechanisms and accuracy; most models perform substantially worse when tested on data from outside their training set. This would not be the case if our collected data were independent and identically distributed samples from the target population. Thus, $\hat{p}$ is a biased estimator of the true model error $\bar{p}$ \cite{lohrSamplingDesignAnalysis2021}. The mean squared error, which is the expected value of the squared error between $\hat{p}$ and $\bar{p}$, is given by the formula 
\begin{equation*}
    \text{MSE}(\hat{p}) = E\left[(\hat{p} - \bar{p})^2\right]
\end{equation*}
which can be written as 
\begin{equation*}
    \text{MSE}(\hat{p}) = E\left[\text{Corr}^{2}(R, p)\right]\times \frac{N-1}{n}\left(1 - \frac{n}{N}\right) \times \sigma_p^2
\end{equation*}
where 
\begin{equation*}
    \text{Corr}(R, p)=\frac{\sum_{i=1}^N (R_{i}-\bar{R})(p_{i}-\bar{p})}{(N-1)S_{R}S_{p}}
\end{equation*}
is the correlation between the sampling mechanism and accuracy. In the formula for $\text{MSE}(\hat{p})$, $n$ is the size of $\hat{T}$, $N$ is the size of the target population, and $\sigma_p^2$ is the population variance of the parameter $p$ \cite{mengStatisticalParadisesParadoxes2018, lohrSamplingDesignAnalysis2021}. We note that that 
\begin{equation*}
    \sigma_{p}^{2}=\bar{p}(1-\bar{p})
\end{equation*}
Since we don't know the value of $\bar{p}$, we follow standard practice and assume that $\sigma_{p}^{2}=\frac{1}{4}$ \cite{glenn1992determining}. For ease of notation, let $\rho_{R,p}:=\text{Corr}(R, p)$. The best bound we can achieve on on $E[\rho_{R,p}^2]$ is 
\begin{equation*}
E[\rho_{R,p}^{2}]\leq\min\left(\frac{n}{N-n}, \frac{N-n}{n}\right)
\end{equation*}
\cite{mengStatisticalParadisesParadoxes2018}. Plugging this back into the formula for the $\text{MSE}$, we see that for $n < \frac{N}{2}$,
\begin{equation*}
    MSE(\hat{p}) \leq \frac{N-1}{N} \cdot \sigma_{p}^{2} \approx \sigma_p^2
\end{equation*}
Thus, the upper bound on our error estimate provides no useful information. Testing on additional samples will not decrease $\text{MSE}(\hat{p})$, since $\text{MSE}(\hat{p})$ has no dependence on $n$.

Following the analysis in \cite{lohrSamplingDesignAnalysis2021}, suppose that $n = \frac{N}{2}$ and $E\left[\rho^2_{R, p}\right] \approx 0.01$. Qualitatively, suppose that we sample half the population and that the correlation between our sampling mechanism and accuracy is relatively small. Then, 
\begin{align*}
    \text{MSE}(\hat{p}) &\approx 0.01 \times \frac{N-1}{n}\left(1 - \frac{n}{N}\right) \times \sigma_{p}^{2} \\
    &\approx\frac{\sigma_{p}^{2}}{100} \\
    &\leq\frac{1}{400} \\
    &=0.25 \%
\end{align*}
This is the same error estimate one would achieve by taking a random sample of $100$ units. By Lyapounov's inequality \cite{billingsleyProbabilityMeasure1995},

\begin{align}
    \text{MAE}(\hat{p}) &= E\Big|\bar{p} - \hat{p}\Big| \\
    &\leq E\Big[\left|\bar{p} - \hat{p}\right|^2\Big]^{1/2}\\
    &= \sqrt{\text{MSE}(\hat{p})} \\
    &\leq \sqrt{\frac{1}{400}} \\
    &= \frac{1}{20} \\
    &= 5\%
\end{align}

where $\text{MAE}(\hat{p})$ is the mean absolute error. Thus, to be confident that $\hat{p}$ is within $5\%$ of $\bar{p}$, we need to sample at least half the population. If $N = 10^{20}$, then we need to sample $5 \times 10^{19}$ images. This is an obscenely large number that no dataset will ever approach in size. Since small amounts of bias in the sampling scheme can have extremely large effects on the quality of an inference \cite{copasInferenceNonrandomSamples1997}, one must sample an extremely large number of images to compensate for sampling biases. 

We make a couple of observations here. Firstly, there are no established rules for determining what constitutes a small correlation between the sampling mechanism and accuracy. A correlation of $0.1$ is considered small in the social sciences \cite{cohenStatisticalPowerAnalysis1988a}, but more work must be done to determine which correlations between the sampling mechanism and accuracy in image datasets are small and which are large.

Secondly, it is impossible to estimate $\rho^2_{R, p}$ from the sampled dataset \cite{mengStatisticalParadisesParadoxes2018}. Thus, any method that calculates $\text{MSE}(\hat{p})$ must rely on an estimate of $\rho^2_{R, p}$. Estimating the value of $\rho^2_{R, p}$ requires knowledge of the true value of $\bar{p}$, or knowledge/assumptions regarding the sampling mechanism, $R$ \cite{mengStatisticalParadisesParadoxes2018}. There would be no need to estimate the value of $\rho^2_{R, p}$ if one already had knowledge of $\bar{p}$. Thus, in all practical situations, we must estimate the value of $\rho^2_{R, p}$ based on our understanding of the sampling mechanism and the target population. 

In domains where we can calculate $\bar{p}$ from a random sample, we can compare the results from non-random samples with random samples to estimate $\rho^2_{R, {p}}$. Using an estimate of $\rho^2_{R, {p}}$ derived from an initial domain in a subsequent domain requires the assumption that the values of $\rho^2_{R, {p}}$ in both domains are nearly equal. Additional research is needed to test this assumption. Given the complexity of components in the image collection process, from image class specifications to search engine algorithms, one should not believe such an assumption without convincing evidence.

Thirdly, empirical results show that many large non-random surveys often have an accuracy equivalent to a very small randomized survey \cite{mengStatisticalParadisesParadoxes2018}. For instance, one non-random sample of $250,000$ people was equivalent to a random sample of $10$ people \cite{bradleyUnrepresentativeBigSurveys2021}. Hence, even if it were the case that the value of $\rho^2_{R, p}$ was nearly equal across many different data-collection mechanisms, it could still be much too large for any sample size to yield a performance estimate with a reasonable uncertainty. This seems especially likely given empirical evidence of dataset bias \cite{torralbaUnbiasedLookDataset2011, buolamwiniGenderShadesIntersectional2018}. Furthermore, in large populations, it would require a miracle to collect a non-random sample that rivals the quality of a random sample \cite{mengStatisticalParadisesParadoxes2018}. We should not bet on miracles when testing high-risk technology that, if unsafely deployed, could cause human fatalities.

Lastly, sampling additional data will not fix our problems. There is no guarantee that sampling more data will improve the quality of our estimate, and trying to make up for poor data quality with data quantity is ill-fated \cite{mengStatisticalParadisesParadoxes2018}. Whether we sample $1$ billion images or $1$ trillion images, both numbers are a tiny fraction of the target population. 

In this section, we have shown that the upper-bound on the uncertainty for our estimate of the model's performance is useless. We also demonstrated that exceptionally large samples are needed to combat sampling bias. Estimating the magnitude of the sample bias cannot be done from a sample alone. Even if it were the case that one had access to a reasonable estimate of the sample bias, empirical results still show that sample biases are rather large. Finally, collecting more data will not improve our cause, since we can never sample more than a tiny fraction of the target population. These initial results suggest that estimates of a model's performance are likely far away from the true performance. 

\section{Inference with Non-Random Samples} \label{non-probability}
There are two primary approaches for handling non-random samples: model based and pseudo-design based \cite{lenau2021methods, changbaowuStatisticalInferenceNonprobability2022}. 

In the model-based approach, a model is fit using the sample data. Often, this model learns to predict the target variable based on a set of auxiliary variables \cite{lenau2021methods}. For instance, suppose we wanted to determine the average value of the variable $y$ over a population. We measured the value of $y$ as well as a set of auxiliary variables $\textbf{x}$ for a non-random sample of units in the population. One simple model might be 
\begin{equation*}
    y_{i} = \mathbf{x}_{i}^{T} \boldsymbol{\beta} + \epsilon_{i}
\end{equation*}
where $\boldsymbol{\beta}$ is a coefficient vector and $\epsilon_{i}$ is random noise \cite{lohrSamplingDesignAnalysis2021}. Then, the values for the non-sampled units can be estimated using the formula
\begin{equation*}
    \hat{y}_{i}= \mathbf{x}_{i}^{T}\mathbf{\hat{B}}
\end{equation*}
where $\mathbf{\hat{B}}$ is an estimator for $\boldsymbol{\beta}$ \cite{lohrSamplingDesignAnalysis2021}. We can compute the estimator $\mathbf{\hat{B}}$ from the sample, but that relies on the assumption that the conditional distribution of $y$ given $\textbf{x}$ is the same for units in the non-random sample and units in the target population \cite{changbaowuStatisticalInferenceNonprobability2022}. We could also compute $\mathbf{\hat{B}}$ with data from a random sample that measured $y$ and $\textbf{x}$ \cite{changbaowuStatisticalInferenceNonprobability2022}. 

Unlike in a random sample, where the probability that each unit is chosen is known in advance, the probability of units being chosen for a non-random sample is unknown. In the pseudo-design based approach, we estimate the probability of inclusion for units in the sample. The weights are often estimated with auxiliary or external information. After weighting, we treat the sample as if it were a random sample \cite{lohrSamplingDesignAnalysis2021}. For instance, in post-stratification \cite{holtPostStratification1979}, the sample is partitioned into $k$ different groups, called post-strata. Suppose there are $N_h$ units from the population and $n_h$ units from the sample in post-strata $h$, where $N_h$ was determined by a population count. Then, we assume that the inclusion probability for units in post-strata $h$ is $\frac{n_{h}}{N_{h}}$.

Weighting classes are most effective when they discriminate in terms of the variable of interest \cite{lynn1996weighting}. In our case, the variable of interest is the model accuracy. Hence, the mean accuracy among different post-strata should differ, and the performance within each post-strata should be nearly uniform since weighting cannot remove bias due to the sampling mechanism within each class \cite{lynn1996weighting}. 

Auxiliary data, such as a census or high-quality random sample, is essential for inference with non-random samples. The accuracy of an inference from a non-random sample depends on the degree to which the following assumptions hold \cite{changbaowuStatisticalInferenceNonprobability2022}:
\begin{enumerate}
    \item The auxiliary data completely determines the sample inclusion probabilities. 
    \item Every unit has a non-zero probability of being included in the non-random sample.
    \item Inclusion probabilities are independent, given the auxiliary data. 
    \item There exists a sample with the relevant auxiliary variables. 
\end{enumerate}
For assumption $1$, it is an open question to determine a set of features which entirely determine the probability that an image is included in the sample, and such a task is certainly daunting. Assumption $2$ certainly does not hold for images collected from the internet. If an image is not on the internet, it has no chance of being included in the dataset. Assumption $3$ is the least important for valid inferences \cite{changbaowuStatisticalInferenceNonprobability2022}. Assumption $4$, however, is where computer vision faces the biggest challenge. As argued earlier, collecting a census of all images in a target domain is impossible. Furthermore, it is impractical to collect a random sample in most domains. Thus, we have no access to high quality auxiliary data which is the critical ingredient needed for inference.

Even after the application of bias-reduction methods, non-random samples are still biased \cite{lohrSamplingDesignAnalysis2021}. Non-random inference rests on a model of the population data and a set of assumptions to support the model. Incorrect model assumptions will yield confidence intervals that are much smaller than the true size \cite{lohrSamplingDesignAnalysis2021}. This problem is especially prevalent in machine learning, where researchers assume that an image dataset with over a billion images must be representative \cite{pavlidisNumberAllPossible2009}.

\section{Reliability Engineering} \label{reliability}
Perhaps we are being too hard on computer vision systems? Are we holding them to a much higher standard than other technical systems? To answer this question, we look at how other fields assess error rates and product reliability. Specifically, we look to the field of reliability engineering \cite{elsayedReliabilityEngineering2021, oconnorPracticalReliabilityEngineering2011}. Many methods exist for demonstrating the reliability of both hardware and software products \cite{luReliabilityDemonstrationTesting2005,zhuSoftwareUnitTest1997}. We analyze a selection of the most popular and applicable reliability demonstration techniques from both of these fields, and illustrate the challenge with applying them to DNNs. 

\subsection{Life Data Analysis}
Life data analysis is an umbrella term used to describe testing mechanisms where a sample of units is subjected to a loading condition for a period of time \cite{luReliabilityDemonstrationTesting2005}. Loading conditions describe the specific conditions under which an object is tested, such as the amount of heat, mechanical stress, electrical load, and corrosion. The time until failure for each unit is recorded and used to infer product safety and reliability. Estimating how long a physical product will operate with life data analysis is comparable to estimating the value of $\bar{p}$, the model's accuracy. In product testing, one generally tests multiple products under similar stresses. For computer vision, one has a single model, but tests that model under different image conditions. In hardware life data testing, a critical assumption is that the sample of products is randomly chosen or nearly randomly chosen \cite{luReliabilityDemonstrationTesting2005, oconnorPracticalReliabilityEngineering2011}. Thus, the equivalent assumption in computer vision would be that the image sample is randomly chosen. As shown earlier, collecting a random sample of images is generally infeasible.

\subsection{The Accelerated Test Method}
If the failure mechanisms of a product are well understood, progressively greater stresses can be applied to a sample of units until they fail. To use accelerated testing, one must precisely understand why a model fails. If the failure mechanism is poorly understood, testing the failure point will provide little information for assessing model reliability \cite{wassermanReliabilityVerificationTesting2002}. Determining a physical product's failure mechanisms is difficult but often feasible with physical modelling. However, determining the failure mechanisms of a computer vision model is uniquely challenging. There are likely numerous failure mechanisms that interact in complex ways, and determining the relationship between stress and reliability is difficult when there are multiple failure mechanisms tested together \cite{wassermanReliabilityVerificationTesting2002}. Furthermore, how could one determine the failure mechanisms of a DNN without testing? As our current testing mechanisms are biased, the errors they uncover may be biased as well. For instance, if one tested a model on images of animals in well lit conditions, the model will fail on a set of images and one perhaps could extrapolate the weak points. However, this testing will not inform you if the model performs poorly on images taken with dark lighting conditions. Thus, the methods surveyed from hardware reliability engineering cannot provide strong evidence for the safety and reliability of a DNN computer vision system.

\subsection{Structural Testing}
Structural testing is typically based on the control-flow and data-flow of a program. Control flow methods test how much of a program's execution graph was covered by tests \cite{zhuSoftwareUnitTest1997}, and data-flow methods analyze the paths between variable definition and variable usage. Similar methods have been developed for neural networks, analyzing test set coverage over individual and collective neuron activations \cite{usmanOverviewStructuralCoverage2023}. Preliminary experiments demonstrated that there is no significant correlation between current deep neural network coverage metrics and model performance or robustness to adversarial attacks \cite{liStructuralCoverageCriteria2019,dongThereLimitedCorrelation2019}. But even if more robust and accurate DNN coverage metrics are developed, they will not necessarily yield better safety guarantees. Under any neuron coverage metric, high coverage implies that many features which the model learned in training were covered in testing as well. But if a test set achieves high neuron coverage, might this just mean that it shares many of the same biases with the training set? How can we be confident that a DNN's neural activation patterns cover all the features that are relevant and present in the target population? Thus, neural coverage methods are unlikely to become a reliable evaluation metric. 

\subsection{Fault-Based Testing}
Fault-based testing estimates the percentage of faults detected by a test set. Methods commonly used to estimate this value include error seeding and mutation testing. In error seeding, one adds intentional errors into a program \cite{goelSoftwareReliabilityModels1985}. The percentage of the intentional errors caught is then used to estimate the total number of errors left in the program. In mutation testing, many similar programs are created, each with a single change from the initial program. The adequacy of a test set is measured by the percentage of mutant programs for which a test case induces a failure. Mutation testing assumes that the  original program is nearly correct and that test cases which induce simple errors, such as those caused by the mutations, will also induce complex errors \cite{tzerposSoftwareEngineeringTesting2015}. 

In error-seeding, it is assumed that the difficulty of detecting the faults introduced is equivalent to the difficulty of detecting intrinsic faults in the program \cite{zhuSoftwareUnitTest1997}. If we could inject specific faults into a DNN, how could we ensure they are equivalent in difficulty to true faults? This is a complicated problem, given the scale of computer vision models and the complexity of the visual world. The assumptions of mutation testing are also unlikely to hold in computer vision. How can we be sure that the original model is almost correct when, as shown earlier, we have no effective method to bound the uncertainty on our accuracy estimates. Furthermore, even if we were confident that our model was almost correct, how could we know whether the errors induced with mutations are equivalent to true errors? Is changing a single weight or a pathway of weights equivalent to not learning a specific feature of the domain? Many challenging questions surround these methods.

Finally, we note that while similar in many regards, fault-based testing and ablation testing have different objectives. Fault-based testing aims to assess the reliability and safety of a model. Ablation testing studies the contribution of different model components to the overall model performance \cite{sheikholeslamiAblationProgrammingMachine2019}. 

\section{Other Methods} \label{other-methods}
\subsection{Out of Distribution (OOD) Detection}
Some work has been done to detect when a testing image differs from the training set distribution \cite{zhangOutDistributionDetection2021, hendrycksBaselineDetectingMisclassified2016}. Using these tools, one can reject real-world input samples that are not from the training data distribution. While these tools may be useful for certain situations, they face many challenges. Firstly, many products cannot safely shutdown when dealing with an OOD input. A self-driving car on a highway cannot pause its perception systems while the images are out of distribution. For systems that can temporarily pause operations, how can we be sure that the out-of-distribution detector works with a specified accuracy? Current OOD detection methods are evaluated empirically \cite{yangGeneralizedOutofDistributionDetection2022}. As demonstrated throughout this paper, empirically evaluating computer vision systems with test sets is ineffective. Certainly, OOD detection  will improve model performance and decrease errors by some margin. However, clear error estimates with small confidence intervals are crucial for safety critical domains with extremely low acceptable error rates. 

\subsection{Out of Distribution Test Sets}
Several papers have explored testing computer vision models on images explicitly varying from the model's training image distribution \cite{hendrycksNaturalAdversarialExamples2021, fangUnbiasedMetricLearning2013, beeryRecognitionTerraIncognita2018}. This is certainly an interesting research direction, but numerous challenges must be handled before one can argue that domain adaptation demonstrates model trustworthiness. For instance, what features should we train a model on? How many distribution shifts and what distribution shifts must the model be able to generalize to for it to demonstrate trustworthiness? Is a random sample of images required to demonstrate that a model has performed adequately well on a new image distribution?

\subsection{Formal Verification}
Current formal verification methods generally provide guarantees against adversarial examples. They prove that the model's output will not change for some amount of variation in the input image \cite{kouvarosFormalVerificationCNNbased2018}. However, current formal verification methods are not even guaranteed to work over the entire target population \cite{houbenInspectUnderstandOvercome2022}. To formally verify that a model works on all images in the target population will require a specification of the target population distribution \cite{wingTrustworthyAI2021}. Given the complexity and size of the target population in computer vision problems, this is an extremely difficult challenge. If better formal verification methods are developed in the future, they will likely just confirm what the never-ending list of adversarial examples \cite{szegedyIntriguingPropertiesNeural2014,moosavi-dezfooliDeepFoolSimpleAccurate2016,moosavi-dezfooliUniversalAdversarialPerturbations2017,papernot2016limitations,zhouHumansCanDecipher2019,hendrycksNaturalAdversarialExamples2021a,serbanAdversarialExamplesObject2020} already prove: that neural networks are not robust.

\section{Limitations}
We acknowledge that opaque data-driven models will not disappear anytime soon. Furthermore, we don't believe that they are useless. There are absolutely domains where safety is not of critical importance, and deep learning methods can be extremely useful. Our argument is limited to domains where one must demonstrate that a model's failure rate is confidently below a specific threshold. Finally, we note that there are multiple ways to define dataset bias. Our work looks at one such definition, and further work is needed to analyze our arguments under alternative definitions.

\section{Other Factors Influencing Trust}
Throughout this paper, we have focused our attention on safety. However, there are other important factors when assessing DNNs, such as fairness, privacy, and security \cite{wingTrustworthyAI2021}. These factors are also critically important and ought to be considered while evaluating a model. Consider the following example. Suppose we determined that a given DNN has an error rate of $1 \pm 0.5 \%$ and we want to use the DNN in a domain which has an acceptable error rate of $5\%$. There exist two groups in the population, $A$ and $B$. Census data shows that $99\%$ of the population belongs to group $A$ and $1\%$ to group $B$. While the DNN error rate is safely under the tolerable rate, the DNN might perform perfectly on individuals in group $A$ and fail on every individual in group $B$. Thus, while the model is technically safe enough, it is certainly not fair. While statistical safety was the primary focus of the paper, we certainly acknowledge the many other important factors which influence trustworthiness and safety.

\section{Conclusion} \label{conclusion}
Alas, can DNNs demonstrate trustworthiness through empirical evaluation? In most domains, they cannot. We have shown that the current techniques for assessing a model with empirical evaluation are not rigorous enough for the minute acceptable error rates needed for real world processes and machines. Furthermore, the surveyed reliability demonstration techniques from other fields did not translate well to demonstrating the reliability of DNNs in computer vision. The recurring issue throughout our study was the non-randomness of image datasets. Bias can taint a non-random sample in imperceptible ways, and the number of biases which can catastrophically sabotage a non-random sample are immense. While dataset creators can attempt to control for a handful of biases, even a slightly biased test set can cause great over-confidence in the performance of a model.

The first contribution of this paper is demonstrating, through non-random sampling theory, that the use of a withheld test set has no theoretical basis for testing computer vision models. There is no guarantee that a model which performs well on a test set will also perform well in the real world. Furthermore, no dataset will ever be large enough to compensate for biases present in the image selection process. We support our claims with both theoretical work in statistics and empirical results from survey sampling and machine learning. Domains where one can collect a random sample are exceptions to our findings. Performance estimates calculated with random samples are accurate and rest on a sound foundation of theory. However, as of the time of writing, most datasets are not randomly collected samples, and it is likely impractical to collect a random sample in many domains. 

The second contribution of this paper is demonstrating the limits of bias-aware dataset collection. Every year, researchers write papers studying bias in machine learning datasets \cite{10.1145/3442188.3445924, 10.1145/3600211.3604691, 10.1145/3514094.3534147}. Our analysis demonstrates that dataset curators cannot remove all bias from a dataset, and controlling for specific features will likely not reduce bias by a substantial amount. Machine learning bias cannot be solved by fixing datasets alone.

Future work is needed to explore the differences between performance estimates calculated with random test sets and non-random test sets using models trained with a non-random sample. Such experiments are required to precisely quantify the amount of bias present in our datasets. 

Finally, we suggest that future performance assessment techniques ought to prioritize understanding a model's decision making and reasoning processes over the model's accuracy. As we have shown, a model's accuracy on a non-random test set overestimates the model's true accuracy. In the many real-world domains with exceptionally low error rates, over-estimating a model's accuracy can be disastrous. Therefore, knowledge of a model's decision-making process is essential for demonstrating the trustworthiness of a model. Until one can understand why a model makes a given decision, the model should not be deployed in a safety-critical domain.

\section{Acknowledgements} \label{acknowledgements}
This research was supported by the Air Force Office of Scientific Research under award number FA9550-22-1-0538 (Computational Cognition and Machine Intelligence, and Cognitive and Computational Neuroscience portfolios); the Canada Research Chairs Program (Grant Number 950-231659); Natural Sciences and Engineering Research Council of Canada (Grant Number RGPIN-2022-04606).

\printbibliography

\end{document}